\documentclass[11pt]{article}

\usepackage{acl}
\usepackage{amsmath}

\usepackage{microtype}
\usepackage{graphicx}	
\usepackage{hyperref}
\usepackage{booktabs} 
\usepackage{verbatim} 
\usepackage{fancyvrb}
\usepackage{float} 
\usepackage{tabularx}
\usepackage{tikz}
\usepackage{pgfplots}
\usepackage{pgfplotstable}

\newenvironment{cherrypick}[4]{
  \noindent\textbf{Model:} \flnm{#1}, \textbf{Word: }\text{#2} \\
  \textbf{Model Response:} \\
  \texttt{#3} \\
  \textbf{Explanation:} #4 \\
}

\newcommand{\flnm}[1]{
  \ifstrequal{#1}{gemini-full}{gemini-1.5-flash}{%
  \ifstrequal{#1}{gemini-small}{gemini-1.5-flash-8b}{%
  \ifstrequal{#1}{gemini-2}{gemini-2.0-flash}{%
  \ifstrequal{#1}{gemini-2l}{gemini-2.0-flash-lite}{%
  \ifstrequal{#1}{gemini-2.5}{gemini-2.5-flash}{%
  \ifstrequal{#1}{gemini-2.5-pro}{gemini-2.5-pro}{%
  \ifstrequal{#1}{4o-mini}{gpt-4o-mini}{%
  \ifstrequal{#1}{4o}{gpt-4o}{%
  \ifstrequal{#1}{4.1}{gpt-4.1}{%
  \ifstrequal{#1}{o3-mini}{o3-mini}{%
  \ifstrequal{#1}{grok}{grok-3-latest}{%
  \ifstrequal{#1}{claude}{claude-sonnet-4-20250514}{%
  \ifstrequal{#1}{llama-70b}{LLaMA-3.3-70B-Instruct}{%
  \ifstrequal{#1}{llama-8b}{meta-llama-3.1-8b-instruct}{%
  \ifstrequal{#1}{llama-4s}{meta-llama-4-scout-17b-16e-instruct}{%
  \ifstrequal{#1}{anonymous}{Apertus-70B-2509}{%
  \ifstrequal{#1}{deepseek-full}{DeepSeek-R1-0528}{%
  \ifstrequal{#1}{gemma-3}{google-gemma-3-12b-it}{%
  \ifstrequal{#1}{gemma-2}{gemma-2-9b-it}{%
  \ifstrequal{#1}{qwen3-big}{qwen3-235b-a22b-04-28}{%
  \ifstrequal{#1}{qwen}{Qwen3-8B}{%
  \ifstrequal{#1}{mistral}{Mistral-Nemo-Instruct-2407}{%
  \ifstrequal{#1}{deepseek}{DeepSeek-R1-Distill-Llama-8B}{%
  \ifstrequal{#1}{bloomz}{bloom-7b1}{%
  \ifstrequal{#1}{bloomz-small}{bloom-560m}{%
  \ifstrequal{#1}{phi}{phi-4}{%
  \ifstrequal{#1}{aya}{aya-expanse-8b}{%
  \ifstrequal{#1}{croissant}{CroissantLLMChat-v0.1}{%
  \textbf{UNKNOWN MODEL}%
}}}}}}}}}}}}}}}}}}}}}}}}}}}}}

\usepackage{fontspec}
\usepackage{polyglossia}

\setotherlanguage{russian}
\setotherlanguage{french}
\setotherlanguage{italian}
\setotherlanguage{hindi}
\setotherlanguage{chinese}
\setotherlanguage{japanese}
\setotherlanguage{arabic}

\newcommand{\ru}[1]{\foreignlanguage{russian}{#1}}

\newcommand{\hi}[1]{\foreignlanguage{hindi}{#1}}

\setotherlanguage{spanish}
\setotherlanguage{portuguese}
\newfontfamily\hindifont{NotoSerifDevanagari-Regular}[
    Path = ./fonts/,
    Extension = .ttf
]
\newfontfamily\cjkfont{NotoSerifSC-Regular}[
    Path = ./fonts/,
    Extension = .ttf
]

\newfontfamily\japanesefont{NotoSerifJP-Regular}[
    Path = ./fonts/,
    Extension = .ttf
]
\newfontfamily\cyrillicfont{NotoSerif-Regular}[
    Path = ./fonts/,
    Extension = .ttf
]

\title{Benchmarking Concept-Spilling Across Languages in LLMs}

\author{Ilia Badanin \\
  EPFL \\
  \texttt{ilia.badanin@epfl.ch} \\\And
  Daniil Dzenhaliou \\
  EPFL \\
  \texttt{daniil.dzenhaliou@epfl.ch} \\\And
  Imanol Schlag \\
  ETH AI Center \\
  \texttt{ischlag@ethz.ch} \\
  }

\begin{document}

\maketitle
\begin{abstract}

Multilingual Large Language Models (LLMs) exhibit remarkable cross-lingual abilities, yet often exhibit a systematic bias toward the representations from other languages, resulting in semantic interference when generating content in non-English languages---a phenomenon we define as \textit{language spilling}.
This paper presents a novel comparative framework for evaluating multilingual semantic robustness by systematically measuring how models handle polysemous words across languages.
Our methodology provides a relative measure of model performance: when required to generate exactly five meanings, both strong and weak models may resort to meanings from dominant languages, but semantically stronger models do so later in the generation sequence, producing more true meanings from the target language before failing, while weaker models resort to dominant-language meanings earlier in the sequence.
We evaluate a diverse set of open and closed multilingual LLMs using a structured meaning generation task across nine languages, employing a carefully curated benchmark of 100 high-polysemy English words.
Our findings reveal significant variation in semantic robustness across both models and languages, providing a principled ranking system for model comparison without requiring definitive causal attribution of error sources.
We contribute both a scalable comparative benchmark for multilingual semantic evaluation and a rigorous validation pipeline---critical tools for developing more linguistically balanced AI systems.
\end{abstract}

\section{Introduction}

\begin{figure}[t]
  \centering
  \includegraphics[width=\linewidth]{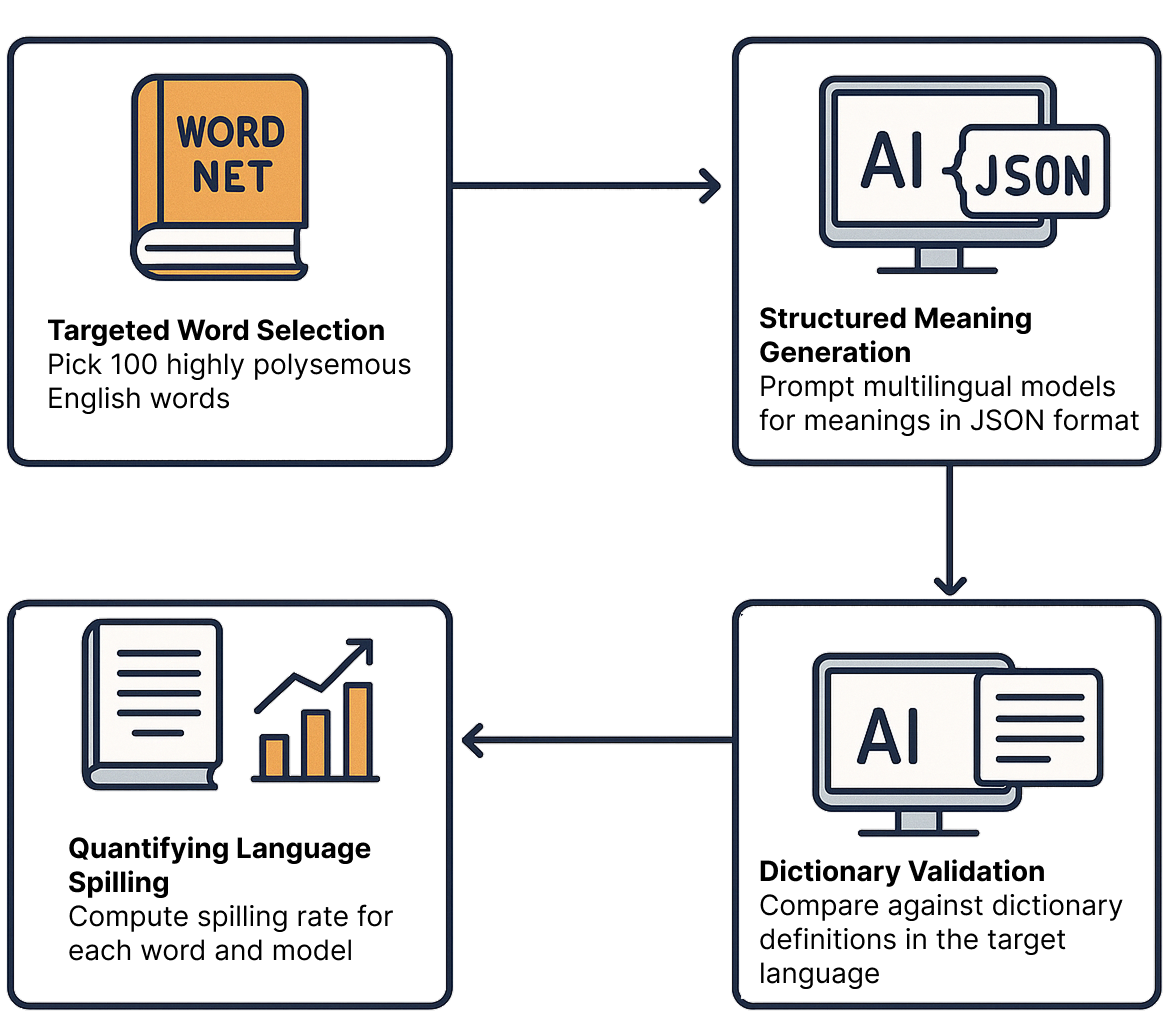}
  \caption{Overview of our methodology: Starting from polysemous word selection, we generate structured meanings across languages, validate them against dictionary references, and compute spilling rates to assess semantic interference.}
  
  \label{fig:overview}
\end{figure}

The advent of Large Language Models (LLMs) has marked a transformative moment in the field of natural language processing, revolutionizing how humans interact with and process information across languages. 
These models, trained on massive text corpora spanning trillions of tokens, demonstrate remarkable capabilities in generating coherent and contextually relevant text in multiple languages \citep{brown2020language, claude3_2024_report, gemini2025thinking_blog}.
Their multilingual abilities have enabled diverse applications including machine translation, cross-lingual information retrieval, content generation, and sophisticated dialogue systems in various languages.
This advancement represents a significant step toward breaking down linguistic barriers in global digital communication and knowledge access.

Despite these advances, a growing body of research suggests that LLMs often exhibit a subtle but pervasive bias toward English \citep{rigouts-terryn-de-lhoneux-2024-exploratory, mihaylov-shtedritski-2024-elegant, wendler2024llamasworkenglishlatent}, the language that typically dominates their training data both in volume and quality. Recent work has shown that multilingual LLMs fundamentally make key decisions in a representation space closest to English, regardless of input language \citep{schut2025multilingual}, and employ an English-pivot multilingual mechanism where non-English queries are converted to English at bottom layers before performing reasoning \citep{huo2025enhancing}. English, due to its dominance on the internet and in various digital resources, typically constitutes the vast majority of the training data. This imbalance has significant implications for how these models learn and represent language. For example, LLaMA-2 was trained primarily on English data, where English tokens volume for training had an 89.70\% \citep{touvron2023llama2openfoundation} share across all tokens.
This English-centric bias manifests in multiple dimensions of language processing: influencing grammatical structures such as word order and agreement patterns \citep{papadimitriou2023multilingual}, stylistic preferences favoring constructions common in English \citep{ahuja2025language}, and most critically, semantic interpretations employed when processing and generating text in non-English languages \citep{cahyawijaya2025towards, liu2023towards}.
The resulting asymmetry in language representation creates fundamental challenges in achieving true multilingual parity in language model performance.

This study focuses on a specific and important aspect of this bias that we term \textit{language spilling}. 
We define language spilling as the phenomenon where multilingual LLMs, when tasked with understanding or generating content in a target language, inadvertently default to semantic representations from other languages—most commonly English due to its dominance in training data, but potentially from any high-resource language with similar representational patterns.

For example, a model might use a French word ``cravate'', which typically means ``necktie'', in contexts where an English writer would use the word ``tie''. This likely happens because of close latent representations of English and French words. We illustrate this phenomenon in Figure~\ref{fig:cravate_example}.

\begin{figure}[t]
\centering
\small
\begin{tabularx}{\linewidth}{lX}
\toprule
\textbf{Word} & \textbf{cravate} (French) \\
\textbf{Core Meaning} & necktie (clothing) \\
\midrule
\textbf{Spilling Example 1} & ``Il a utilisé une cravate pour tirer la selle du cheval'' \\
\textit{Literal Trans.} & ``He used a necktie to pull the horse’s saddle'' \\
\textit{Intended Meaning} & He used a \textbf{rope/tie} to pull... (English interference: tie $\rightarrow$ cravate) \\
\midrule
\textbf{Spilling Example 2} & ``Les arbres étaient liés avec des cravates'' \\
\textit{Literal Trans.} & ``The trees were tied with neckties'' \\
\textit{Intended Meaning} & The trees were tied together with \textbf{ties}. (English interference: tie $\rightarrow$ cravate) \\
\bottomrule
\end{tabularx}
\caption{Example of Language Spilling: The model uses the French word ``cravate'' (necktie) in contexts appropriate for the English word ``tie'' (fasten/bind), resulting in semantically unnatural sentences.}
\label{fig:cravate_example}
\end{figure}

This phenomenon, also observed in related studies \citep{li2025lostliteralismsupervisedtraining, wang2023understandingtranslationesecrosslingualsummarization}, results in outputs that may appear grammatically correct but contain subtle semantic errors that native speakers immediately recognize as unnatural or incorrect. Additional examples across nine languages are provided in Appendix~\ref{sec:apx_qualit}.

The prevalence of language spilling undermines the reliability of multilingual LLMs in real-world applications.
When models fail to capture language-specific semantics and cultural context, they risk misinterpreting user intent and generating misleading content \citep{veselovsky2025localizedculturalknowledgeconserved}. 
This is particularly concerning in high-stakes domains such as legal translation, medical information retrieval, and cross-cultural communication, where semantic precision is essential. 
Related work has demonstrated that LLMs exhibit significant inconsistencies in generating text in intended languages \citep{marchisio-etal-2024-understanding} and produce more unsafe responses for non-English queries than English ones \citep{wang2023all}. 
As LLMs increasingly serve as interfaces between languages, their ability to maintain semantic fidelity across linguistic boundaries becomes crucial to ensure equitable access to AI technologies worldwide.

In this paper, we present a novel comparative methodology to systematically evaluate multilingual semantic robustness in LLMs.
Our approach leverages English words with multiple meanings (high-polysemy words) to assess how well models handle semantic complexity across target languages and understand why polysemic challenges require eliciting stronger multilingual representations.
We instruct models to generate structured meanings and examples for translated terms, then validate these outputs against authoritative target language dictionaries using a judge model with dictionary context.
Critically, our methodology provides a relative measure of semantic robustness: when forced to generate exactly five meanings, both strong and weak models may resort to meanings from dominant languages, but stronger models do so later in the generation sequence, producing a higher number of valid meanings before defaulting to dominant-language semantics, while weaker models resort to them earlier---regardless of whether these errors stem from interference from English, other high-resource languages, hallucinations, or other failure modes.

Through this method, we establish comparative rankings across different languages and models, providing a principled framework for multilingual model selection and development without requiring definitive causal attribution of semantic errors.

Our work makes three key contributions: \textbf{(1)} a novel comparative benchmark for evaluating multilingual semantic robustness across nine diverse languages and multiple models, providing principled rankings for model selection; \textbf{(2)} a rigorous methodology for stress-testing semantic validity through structured meaning generation tasks, validated through extensive human evaluation achieving 77.43\% agreement with native speakers; and \textbf{(3)} comprehensive empirical analysis demonstrating significant variation in semantic robustness across models and languages, with consistent rankings verified through judge model concordance (Kendall's W = 0.9176). These contributions provide essential tools for developing more linguistically balanced and culturally inclusive AI systems.

\section{Method}
\label{sec:method}

Our methodology, illustrated in Figure~\ref{fig:overview}, systematically quantifies semantic robustness in multilingual LLMs through four stages: 
(1) selecting the top 100 high-polysemy English words as probes for semantic probes; 
(2) prompting models to generate five distinct meanings with examples for each translated word across target languages; 
(3) validating responses using a judge model with dictionary definitions to determine semantic validity; and 
(4) calculating spilling rates as the proportion of invalid meanings. This approach provides a comparative framework for evaluating how well models maintain semantic validity. Our validation experiments confirm that this metric captures meaningful semantic interference patterns (Section~\ref{sec:ablation}). 

\subsection{Polysemous Word Selection and Translation}
We identified English words susceptible to semantic misinterpretation by leveraging WordNet's lexical database \citep{miller-1994-wordnet} to select 100 words with the highest number of different meanings. 
These high-polysemy words serve as effective probes for language spilling, as their multiple meanings increase the likelihood of semantic confusion across languages. 
We used GPT-4o \citep{openai2024gpt4ocard} to translate these words into target languages. For example, the English word ``tie'' (with meanings including necktie, knot, draw in a game, etc.) was translated to French as ``cravate''. This translation choice is deliberate: GPT-4o selects translations according to the model's own understanding of meaning, helping us find translated words that stay closely related to the original English terms, making it easier to detect when models inappropriately transfer semantics across languages. We specifically instructed the translation model to avoid Japanese katakana, as katakana usage for foreign words represents an intended borrowing phenomenon that would confound our analysis.

\subsection{Multilingual Meaning Generation}

We implemented a structured JSON-based prompting methodology requiring models to generate five distinct meanings with three examples per meaning for each translated word. 
This ``5 meanings'' format represents a controlled stress test designed to reveal when models begin fabricating content or resorting to representations from other languages. This standardized challenge creates comparable conditions across models and languages, though we acknowledge it may not reflect natural semantic distributions.
The prompts were translated into all target languages to minimize bias from English, ensuring that both the prompt and the target words were presented in the same language.
Our comparative analysis in Section~\ref{sec:json_no_json} confirmed that while structured formatting did not affect spilling rates, it significantly reduced the noise of formatting inconsistencies and facilitated downstream processing. 
The complete prompt template for meanings generation is available in the Appendix~\ref{sec:apx_gen_prompt}.

\subsection{Dictionary-Based Validation}

To check if the generated meanings were accurate in the target languages, we used a more powerful model as a judge, incorporating relevant dictionary definitions in its context for each word in the target language to make more informed decisions. 
This validation process objectively evaluates the generated meanings by leveraging dictionary entries from authoritative sources as the ground truth. 
For each target language, we carefully selected and integrated comprehensive dictionary excerpts from reliable sources to provide a robust reference framework (see Appendix~\ref{sec:dictionary_sources} for complete details on dictionary sources and validation methodology). Our final dictionary excerpts were reviewed and refined by native speakers. To ensure the reliability of our automated approach, we conducted extensive human evaluation with native speakers, achieving strong agreement between human annotators and our judge model.

The judge prompt instructs the model to pay attention to the information provided from the dictionary and return a set of five \texttt{True/False} values for each of the definitions generated by the smaller model. 
This approach transforms the complex task of determining whether a generated meaning is a valid sense into a simplified matching operation against established lexical references. 
The complete prompt template for meanings validation is available in the Appendix~\ref{sec:apx_judge_prompt}.

By aggregating the judge model's assessments across words, models, and languages, we calculate a \textit{spilling rate} for each instance, representing the percentage of meanings determined to be invalid. This quantitative measure reflects how often a model produces incorrect or English-influenced meanings in other languages, providing a robust metric to compare different models in terms of multilingual comprehension.

Please, refer to Section~\ref{sec:abl_judges} for more details on how the choice of prompt and judge affects the alignment with human preferences.

\subsection{Spilling Rate Quantification}

Our methodology provides a relative measure: when forced to generate exactly five meanings, both strong and weak models may resort to meanings from dominant languages, but semantically stronger models do so later, producing more true meanings from the target language before resorting to dominant-language semantics. For example, on average llama-3.1-8b-instruct shows 41\% spilling rate while Apertus-70B-2509 shows 20\%, meaning the former exhausts valid semantic content much faster.
For each word-language-model combination, we calculate this rate as the percentage of meanings deemed invalid by the judge model when compared against authoritative dictionary definitions. 
A meaning is considered invalid when that meaning does not appear in dictionary entries for the corresponding word---this rejection may result from English semantic interference, hallucinations, or other failure modes, but the key insight is that stronger models should be more robust across all error types.
Our validation experiment in Section~\ref{sec:abl_eng_exp} confirms that approximately 70\% of dictionary-rejected meanings in foreign languages are indeed supported by English dictionaries, validating the presence of cross-lingual interference while acknowledging other error sources.
This binary classification approach (valid/invalid) enables principled model comparison at multiple levels: per word (revealing which concepts challenge models most), per language, and per model (establishing semantic robustness rankings). The spilling rate functions as a comparative ranking system rather than an absolute measure of English interference. 
To ensure measurement consistency, we discard any responses where the judge model returns non-standard outputs or where the generative model fails to produce five distinct meanings in the required format. This filtering mechanism maintains data quality while providing statistically robust comparisons across the evaluation dimensions.

\section{Results}
\label{sec:results}

This section presents the key findings of our evaluation of language spilling in multilingual LLMs using the methodology described in the previous section. 
Our analysis focuses on quantifying the average performance of the generative models and the detailed performance across different target languages. 
The impact of different factors on the evaluation results is explored in Section \ref{sec:ablation}.

\subsection{Models and Languages}
In this work, we analyze nine languages --- Russian, French, Chinese, Japanese, Italian, German, Portuguese, Spanish, and Hindi.
We use 16 different models for evaluation from 3 different classes: small open-source models, big open-source models, and closed-sourced models. Please, refer to Appendix~\ref{sec:apx_models} to find detailed information about the models.
We used Gemini 2.5-Flash \citep{gemini25_technical_report} as the main judging model. The impact of choice of a judge model is covered in Section~\ref{sec:abl_judges}.

\subsection{Average Performance of Generative Models}

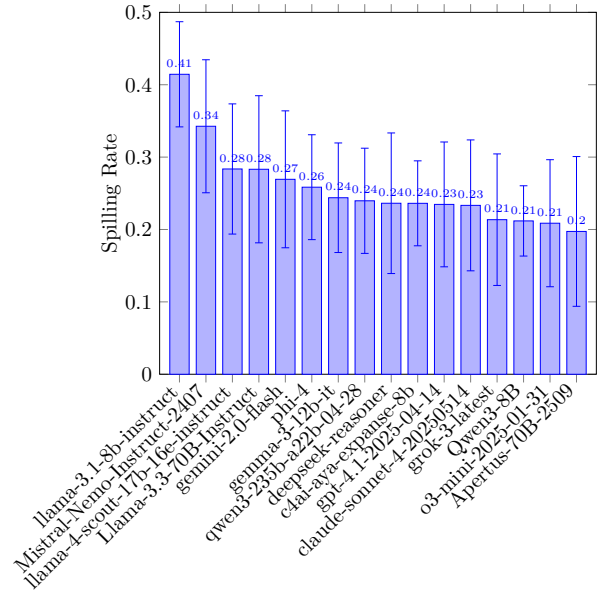
\begin{figure}
\resizebox{0.5\textwidth}{!}{
\begin{tikzpicture}
\begin{axis}[
    ybar,
    ymin=0,
    ymax=0.5,
    width=9.5cm,
    bar width=10pt,
    xtick=data,
    xticklabel style={rotate=45, anchor=east},
    symbolic x coords={
llama-3.1-8b-instruct, Mistral-Nemo-Instruct-2407, llama-4-scout-17b-16e-instruct, Llama-3.3-70B-Instruct, gemini-2.0-flash, phi-4, gemma-3-12b-it, qwen3-235b-a22b-04-28, deepseek-reasoner, c4ai-aya-expanse-8b, gpt-4.1-2025-04-14, claude-sonnet-4-20250514, grok-3-latest, Qwen3-8B, o3-mini-2025-01-31, Apertus-70B-2509
    },
    ylabel style={yshift=-10pt},
    ylabel=Spilling Rate,
    enlarge x limits=0.05,
    nodes near coords,
    nodes near coords style={font=\tiny},
    every node near coord/.append style={
        /pgf/number format/.cd,
            fixed,
            precision=2
    },
]
\addplot+[
    error bars/.cd,
        y dir=both,
        y explicit,
] table[
    x=Model,
    y=Spilling Rate,
    y error=std,
    col sep=comma
] {data/model_rates.csv};
\end{axis}
\end{tikzpicture}
}
    \caption{Average Spilling Rate Across Generative Models. Lower is better.}
    \label{fig:average_model_performance_results}
   
\end{figure}

Figure \ref{fig:average_model_performance_results} illustrates the average language spilling rate observed across all target languages for each of the generative models evaluated in this study. 
This provides a high-level comparison of how different models tend to default to English-based representations when generating meanings for polysemous words in other languages. 
As shown in Figure \ref{fig:average_model_performance_results}, model llama-3.1-8b-instruct exhibited the highest average spilling rate at 41\%, while model Apertus-70B-2509 demonstrated the lowest at 20\%.

\newcommand{\minitext}{\fontsize{1pt}{1pt}\selectfont}

\subsection{Average Language Performance}

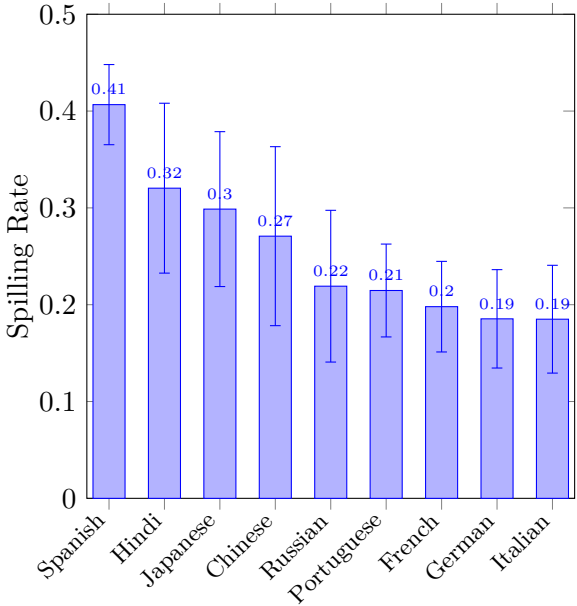
\begin{figure}
\begin{tikzpicture}
\begin{axis}[
    ybar,
    ymin=0,
    ymax=0.5,
    width=8cm,
    height=8cm,
    bar width=12pt,
        xtick=data,
    symbolic x coords={
Spanish, Hindi, Japanese, Chinese, Russian, Portuguese, French, German, Italian
    },
    xticklabel style={rotate=45, anchor=east, font=\footnotesize},
    x tick label style={font=\footnotesize},
    ylabel style={yshift=-10pt},
    ylabel=Spilling Rate,
    nodes near coords,
    every node near coord/.append style={font=\tiny, rotate=0},
    enlarge x limits=0.05,
    axis background/.style={fill=white},
    bar shift=0pt,
]
\addplot+[
    error bars/.cd,
        y dir=both,
        y explicit,
] table[
    x=Language,
    y=Spilling Rate,
    y error=std,
    col sep=comma
] {data/language_rates.csv};

\end{axis}
\end{tikzpicture}
\caption{Average Spilling Rate Across Languages. Lower is better.}
\label{fig:average_language_performance_results}
\end{figure}

Figure \ref{fig:average_language_performance_results} summarizes the average language spilling rate observed for each of the nine target languages, aggregated across all generative models. This perspective highlights which languages, on average, are more susceptible to eliciting English-based semantic defaults in the LLMs under investigation. 
Interestingly, we observed a significant disparity among Romance languages: while French (20\%), Portuguese (21\%), and Italian (19\%) show nearly identical low spilling rates, Spanish is a distinct outlier. We hypothesize this could reflect: (1) differential training data quality and quantity, (2) potential differences in the degree of English semantic borrowing in modern Spanish usage compared to other Romance languages, or (3) variations in our dictionary coverage quality across these languages.
Table~\ref{tab:agg_spilling_rates} reveals significant variations in spilling rates across languages. 
For instance, model DeepSeek-R1-0528 showed a particularly high spilling rate for Spanish (47\%), while its performance on German was considerably better (16\%). Generally, for most models German appears to be a language with the least spilling rate.
To examine a more granular view of the phenomenon of language spilling, please refer to the Appendix~\ref{sec:apx_pics}

\pgfplotstableread[col sep=comma]{data/sorted_language_model_performance.csv}\datatable


\def\mycols{German,Spanish,French,Hindi,Italian,Japanese,Portuguese,Russian,Chinese}

\pgfplotstableset{
    highlight max row/.style={
        postproc cell content/.code={
            \begingroup
            \pgfmathsetmacro{\maxval}{-1e100} 
            \foreach \col in \mycols {
                \pgfplotstablegetelem{\pgfplotstablerow}{\col}\of{\datatable}
                \pgfmathparse{\pgfplotsretval}
                \pgfmathsetmacro{\maxval}{max(\maxval, \pgfmathresult)}
            }
\pgfplotstablegetelem{\pgfplotstablerow}{\pgfplotstablecolname}\of{\datatable}
            \let\currentcellval=\pgfplotsretval
            \pgfmathfloatcompare{\currentcellval}{\maxval}
            \ifnum\pgfmathresult=0 
                \pgfkeysalso{@cell content/.add={\bfseries}{}}
            \fi
            \endgroup
        }
    }
}

\begin{table*}[ht!]
    \centering
    \caption{Aggregated Language Spilling Rate (\%) by Generative Model and Language. Lower is better.}
    \label{tab:agg_spilling_rates}
    \resizebox{1\textwidth}{!}{
        \pgfplotstabletypeset[
            col sep=comma,
            columns/Model/.style={string type, column name=\textbf{Generative Model}},
            columns/Spanish/.style={fixed, precision=1, multiply with=100, column name=\textbf{Spanish}},
                        columns/German/.style={fixed, precision=1, multiply with=100, column name=\textbf{German}},
            columns/French/.style={fixed, precision=1, multiply with=100, column name=\textbf{French}},
            columns/Hindi/.style={fixed, precision=1, multiply with=100, column name=\textbf{Hindi}},
            columns/Italian/.style={fixed, precision=1, multiply with=100, column name=\textbf{Italian}},
            columns/Japanese/.style={fixed, precision=1, multiply with=100, column name=\textbf{Japanese}},
            columns/Portuguese/.style={fixed, precision=1, multiply with=100, column name=\textbf{Portuguese}},
            columns/Russian/.style={fixed, precision=1, multiply with=100, column name=\textbf{Russian}},
            columns/Chinese/.style={fixed, precision=1, multiply with=100, column name=\textbf{Chinese}},
            every head row/.style={before row=\toprule, after row=\midrule},
            every last row/.style={after row=\bottomrule},
        ]{\datatable}
    }
\end{table*}

\section{Ablation Study}
\label{sec:ablation}

We conducted several ablation studies to evaluate the robustness and performance of our methodology. These experiments focused on the impact of the generation format and the verification process.



\subsection{Impact of Generation Format (JSON)}
\label{sec:json_no_json}

We tested prompting with and without JSON formatting to assess its influence on eliciting semantic spillings and the ease of extracting distinct meanings. 

We evaluated the top 15 Russian words manually using the LLaMA-3.3-70B model under the same conditions as in our automatic quantification.
The results showed identical spilling rates of 54.67\% (41 out of 75 produced meanings) for both JSON and non-JSON formats.
The model demonstrated the same exact spilling cases and the same exact correct usages of a word in both conditions, essentially replicating its generation patterns while adding syntax symbols for the JSON case.

Despite the identical spilling rates, we opted to use JSON formatting for three key reasons: (1) it kept responses shorter and more structured, (2) it made downstream parsing significantly easier, and (3) it facilitated the judge model's ability to read and evaluate the generated text.

\subsection{Impact of Judge Model Variation}
\label{sec:abl_judges}

To assess the potential influence of the choice of the evaluation model on the measured spilling rates, we conducted an ablation study comparing the average spilling rates obtained when using different judge models and prompts. 

\begin{table}[ht!]
    \centering
    \caption{Model Alignment with User Preferences}
    \label{tab:model_performance_comparison}
    \resizebox{0.5\textwidth}{!}{
        \begin{tabular}{lccccc}
            \toprule
            \textbf{Prompt Index} & \textbf{DeepSeek-R1-0528} & \textbf{gemini-2.5-flash} & \textbf{gemini-2.5-pro} & \textbf{o3-mini} \\
            \midrule
            1 & 0.7340 & 0.6964 & --     & 0.7036 \\
            2 & 0.7407 & 0.7298 & 0.7395 & 0.7564 \\
            3 & 0.7415 & 0.7523 & 0.7560 & 0.7321 \\
            4 & 0.7352 & 0.7456 & 0.7341 & 0.7209 \\
            5 & 0.7389 & 0.7491 & 0.7395 & 0.7561 \\
            6 & 0.6952 & 0.6338 & 0.7184 & 0.7250 \\
            7 & 0.7467 & \textbf{0.7743} & 0.7589 & 0.7564 \\
            8 & 0.7578 & 0.7468 & 0.7395 & \textbf{0.7615} \\
            9 & 0.7560 & 0.7450 & 0.7486 & 0.7596 \\
            10 & 0.7519 & 0.7371 & \textbf{0.7743} & 0.7376 \\
            11 & \textbf{0.7744} & 0.7505 & 0.7660 & 0.7106 \\
            \midrule
            \textbf{Min} & 0.6952 & 0.6338 & 0.7184 & 0.7036 \\
            \textbf{Max} & \textbf{0.7744} & \textbf{0.7743} & \textbf{0.7743} & \textbf{0.7615} \\
            \bottomrule
        \end{tabular}
    }
\end{table}

Initially, we explored automated prompt optimization using TextGrad \citep{yuksekgonul2024textgradautomaticdifferentiationtext}, but found that it did not improve performance on our specific validation task. Instead, we manually designed the 11 alternative prompts for semantic judgment. 

We tested 4 high-performing models (DeepSeek-R1-0528, Gemini-2.5-Flash and Gemini-2.5-Pro, and OpenAI's o3-mini) with 11 different prompts each (Table~\ref{tab:model_performance_comparison}), identifying the best prompt for each judge model. 
The human agreement scores were as follows: DeepSeek 77.44\%, Gemini 77.43\%, Gemini 2.5 Pro 76.60\%, and o3-mini 76.15\%.
In the same subset, the pairwise agreement between the models ranges from 84.59\% to 92.61\%, which is a strong indicator of the similarity of the models' judgement. 
Regardless of the specific judge model used, judges order generative models almost identically, as evidenced by Kendall's coefficient of concordance (W = 0.9176), indicating very strong agreement among judges. 
In our final experiments, we ultimately chose Gemini Flash 2.5 as our primary judge model with the prompt No7.

\subsection{Language Spilling Validation Experiment}
To validate that our metric captures actual language spilling rather than general semantic errors, we conducted an additional experiment using LLaMA models, which are known to be heavily English-biased, since 89.70\% of the training data consisted of English \citep{touvron2023llama2openfoundation}. 

Using meaning generations for 100 words and in 9 languages, we filtered for foreign language meanings that were rejected by the foreign dictionary according to our judge. 
The incorrect meanings were then translated into English and evaluated using the same judge with an English dictionary (WordReference). 
In this setting, we found that approximately 70\% of the meanings that were rejected in the foreign language were accepted in the English language. 
This finding indicates that the majority of invalid meanings are not random hallucinations but rather instances of cross-lingual semantic transfer. While this does not definitively prove causation at the representational level, the high proportion of English-valid meanings (70\%) provides strong empirical support for the hypothesis that English semantic interference is a dominant source of error in multilingual generation.

\section{Related Work}

\subsection{English Bias in Multilingual LLMs}

Multilingual LLMs often inherit English-centric semantic biases, producing outputs in other languages that reflect English grammar or word senses. 
In \citet{papadimitriou2023multilingual} authors first documented ``grammatical structure bias'', demonstrating that multilingual BERT preferred English-like syntax (explicit pronouns, SVO word order) even when modeling Spanish or Greek. 
This work shows that higher-resource language structures ``bleed into'' lower-resource ones, a phenomenon similar to the \textit{language spilling} we investigate. Similarly \citet{chen2025crosslingual} 
note that most LLMs are trained on English corpora and ``are not able to replicate similar success in other languages''. 
In machine translation \citet{zhang-toral-2019-effect} find that translationese (texts translated from English) are systematically easier for MT systems. 
A translationese-free evaluation protocol \citet{ahuja2025language} specifically compare English versus native instructions for multilingual LLMs, finding that models still ``adhere more closely to English instructions'' even on non-English tasks. 
Together, these works document an English-default bias in grammar and prompt-following across multilingual models.

\subsection{Cross-Lingual Polysemy and Meaning Preservation}

A major facet of \textit{language spilling} is polysemy – using an English-based sense where the target language would use a different sense. 
A cross-lingual sense-disambiguation benchmark StingrayBench \citep{cahyawijaya2025towards} is built on false friends (words that look alike but differ in meaning) show that LLMs often pick the ``higher-resource'' sense. 
For example, ``Angel'' has a false friend in German where the same writing means a ``fishing rod''. 
They find that LLM outputs tend to be biased toward the high-resource language's meaning. 
\citet{liu2023towards} similarly study ambiguous sentences with highly polysemous English words. 
They translate such sentences into Spanish, Italian, German, Russian, and Chinese and ask LLMs to disambiguate. 
Strikingly, very large LLMs (e.g., BLOOMZ-176B \citep{muennighoff2022crosslingual}, LLaMA-65B \citep{touvron2023llamaopenefficientfoundation})  matched or outperformed top MT systems, resolving ambiguous senses correctly most of the time. However, \citet{campolungo-etal-2022-dibimt} found that conventional NMT systems fail on 40–50\% of these cases due to entrenched biases in training data. These works highlight that multilingual models struggle to accurately preserve meaning across languages when faced with polysemy, often defaulting to an English-centric interpretation of word senses.

\subsection{Structured Prompting and Error Diagnosis}
\label{sec:abl_eng_exp}

Finally, several methods use structured prompting to force LLMs into explicit semantics and detect errors. 
For example, ASPIRO \citep{vejvar2023aspiro} uses a JSON-format output prompt to verbalize RDF triples, enforcing name-entity agnostic templates and including parser-based error checks. By repeating generation until the output parses correctly, it significantly reduces generation errors. This illustrates how error-inducing structured prompts can steer an LLM to produce the intended semantics. 
In a similar spirit, MLPrompt \citep{li2025large} translates error-prone constraints into a non-dominant language to help catch mistakes. Since LLMs often overlook rules stated in a dominant language like English, rewriting a violated rule in, say, Spanish, can strengthen the model’s understanding and reasoning.

Such cross-lingual prompting can expose English-default assumptions: the model is forced to engage with the semantics of another language when verifying its output. 
Chain-of-Dictionary prompting \citep{lu2024chainofdictionarypromptingelicitstranslation} likewise adds multilingual dictionary chains into prompts to improve translation by explicitly including lexical semantics. 

\section{Limitations}
This study faces several inherent limitations, primarily rooted in the challenges of evaluating semantic spillings across different languages. While the methodology simplifies the process through automation, the accuracy of the approach depends on having a carefully designed set of prompts and words that are likely to induce spillings. 

An additional limitation stems from the reliance on dictionary definitions as the sole source of ground truth. Dictionaries may not always capture the full spectrum of a word's usage, including idiomatic expressions or culturally specific connotations, which could lead to instances being classified as spillings when they represent valid but less common interpretations. Furthermore, our analysis primarily focused on polysemous words. Semantic spilling might manifest differently with other linguistic phenomena such as idiomatic expressions, metaphors, or words with strong cultural connotations.

\section{Future Work}

In our extended work, we aim to collaborate with linguistics departments to refine and expand the dictionary resources used in our evaluation pipeline. Additionally, we plan to incorporate human evaluations to more directly assess the accuracy and nuance of the model-generated meanings. Future work should also incorporate sense-aligned multilingual dictionaries to better handle translation ambiguity, as sense distinctions differ across languages and our current approach uses single default translations. Furthermore, we plan to extend our evaluation to context-dependent usage (pragmatic competence), examining spilling behavior when words are presented within sentences rather than in isolation.

Beyond English-centric spilling, we will also explore cross-language spilling between closely related or script-sharing languages---such as Russian to Bulgarian, Spanish to Basque, or Hindi to Sanskrit. These cases are particularly interesting because shared scripts or token overlap may cause models to incorrectly assume semantic similarity, leading to subtle but systematic errors. We expect that these forms of structural or token-level interference will reveal further challenges in multilingual semantic representation.

\section{Conclusion}

This study introduced a novel methodology for systematically investigating and quantifying the phenomenon of language spilling in multilingual Large Language Models. By employing a structured approach involving the generation of JSON-formatted meanings for translated polysemous English words, automated validation against target language dictionaries using a larger LLM with dictionary inclusion, and comprehensive human evaluation for validation, we were able to identify and analyze instances where models inadvertently defaulted to English-based semantic representations.

Our findings revealed significant variations in language spilling rates across different generative models and target languages, highlighting the nuanced challenges these models face in truly mastering multilingual semantics. The ablation studies confirmed that JSON formatting did not affect spilling rates but improved parsing and readability for the judge LLM. The exploration of different judge models in the ablation study provided insights into the stability of our evaluation framework.

Ultimately, this research contributes to a deeper understanding of the complexities inherent in multilingual LLMs and provides a foundation for future work aimed at enhancing their cross-lingual semantic accuracy and reliability, paving the way for more robust and trustworthy multilingual natural language processing systems. To ensure reproducibility and support future research, we will make our complete codebase, benchmark datasets, dictionary resources, and human evaluation annotations publicly available.



\bibliography{custom}

\newpage
\appendix   

\onecolumn

\section{Prompts}
\subsection{Meanings Generation Prompt}
\label{sec:apx_gen_prompt}

\raggedbottom

\texttt{
Create a list of 5 distinct meanings for the word '\%s'. For each meaning, provide exactly 3 examples of its usage. Your response must be in JSON format only. Do not include anything else in your response. The JSON should follow this structure: { meaning: [examples], meaning: [examples], ... }
\\
}

\subsection{Translation Generation Prompt}
\label{sec:apx_translation_prompt}

The following prompt was used to generate translations for the polysemous word set:

\begin{verbatim}
You are a multilingual translator. Given a list of English words, translate each word 
into: **Russian (ru)**, **French (fr)**, ..., and **Hindi (hi)**. Ensure that 
translations are **nouns whenever possible**. Ensure that Japanese translations are 
**not** in katakana. Return the translations as a JSON object in the following format: 
{"translations": {"word1": {"ru": "Russian translation", "fr": "French translation", 
..., "hi": "Hindi translation"}, "word2": {"ru": "Russian translation", ...} ...}}
Here is the list of words to translate: {words_batch}
\end{verbatim}

\subsection{Judging LLM Prompt}
\label{sec:apx_judge_prompt}

We structured the prompt using XML tags to enhance response quality, as recommended in their prompt design guidelines \citep{google2024structuredprompts}.

\begin{verbatim}
You are an AI assistant specializing in verifying meanings of terms using a structured 
knowledge base. Your primary task is to evaluate whether the meanings provided in the  
section match, partially align with, or differ from the ground truth meanings stored in 
the  section. The  section may contain unstructured text, and your goal is to extract 
relevant terms and assess their meanings accurately.
<DICTIONARY>
%s
</DICTIONARY>
<INSTRUCTIONS>
0. Knowledge to match to is DICTIONARY
1. Queries that may not be structured
2. List of `true/false` values for each query. No other output
3. Matching Logic:
   - Check if the 'query' matches any 'knowledge entry'. Return 'true' if match found, 
   'false' 
   if not.
   - The number of output entries should match the number of query entries
   - It should be true if it's close
   - The response should be as a JavaScript list [], with true, false
</INSTRUCTIONS>
<QUERY>
%s
</QUERY> 
\end{verbatim}


\section{Models}
\label{sec:apx_models}

\begin{table}[H]
\centering
\begin{tabularx}{\textwidth}{|X|c|c|}
\hline
\textbf{Model Name} & \textbf{Type} & \textbf{\# Parameters} \\
\hline
\flnm{qwen}\citep{yang2025qwen3technicalreport} & Open Source & 8B \\
\flnm{llama-8b}\citep{grattafiori2024llama3herdmodels} & Open Source & 8B \\
\flnm{aya}\citep{dang2024ayaexpansecombiningresearch} & Open Source & 8B \\
\flnm{mistral}\citep{mistral2024nemo} & Open Source & 12B \\
\flnm{gemma-3}\citep{gemmateam2025gemma3technicalreport} & Open Source & 12B\\
\flnm{phi}\citep{abdin2024phi4technicalreport} & Open Source & 14B \\
\hline
\flnm{llama-70b}\citep{grattafiori2024llama3herdmodels} & Open Source & 70B \\
\flnm{anonymous}\citep{hernándezcano2025apertusdemocratizingopencompliant} & Open Source & 70B \\
\flnm{llama-4s}\citep{meta2025llama4scout} & Open Source & 109B \\
\flnm{qwen3-big}\citep{yang2025qwen3technicalreport} & Open Source & 235B \\
\flnm{deepseek-full} \citep{deepseekai2025deepseekr1incentivizingreasoningcapability} & Open Source & 671B \\
\hline
\flnm{grok}\citep{xai2025grok3} & Closed Source & Unknown \\
\flnm{4.1}\citep{openai2025gpt41} & Closed Source & Unknown \\
\flnm{o3-mini}\citep{openai2025o3mini} & Closed Source & Unknown \\
\flnm{claude}\citep{anthropic2025claudeSonnet4} & Closed Source & Unknown \\
\flnm{gemini-2}\citep{google2024gemini2flash} & Closed Source & Unknown \\
\hline
\end{tabularx}
\caption{Evaluated models with type and parameter count.}
\label{tab:models}
\end{table}

Among the evaluated models, only LLaMA, Aya, and Apertus explicitly disclose training on multilingual data and languages used for training. 
Although the LLaMA-2 family was not included in this exploration, it is worth noting that it remains one of the few popular models that has disclosed detailed statistics on language distribution in its training data.

For the remaining models, although they demonstrate capabilities in multilingual tasks and are frequently evaluated on multilingual benchmarks, no information is provided regarding the actual linguistic composition of their training data. As such, we cannot say what languages these models were exposed to during pretraining.

This lack of transparency limits interpretability when comparing cross-lingual behavior and bias, particularly with respect to phenomena like language spilling. With language distribution data for just one model, drawing rigorous correlations with training data composition is not statistically sound. That is why we are focusing on linguistic and architectural factors we can reliably analyze across all models.

OpenAI, DeepSeek, Cohere's Aya, Gemini, Grok, and Claude models were accessed through their official APIs. Qwen3-235B, LLaMA-4 Scout, Gemma 3, and Mistral Nemo were used via the OpenRouter API. Apertus-70B-2509, LLaMA-3.3-70B, Qwen3-8B, and Phi-4 were launched on a local cluster. No additional generation parameters were passed; all models used default generation and tokenizer parameters. API usage ensures compliance with respective terms of service and does not violate any terms of agreement.

\section{Supplementary Figures}
\label{sec:apx_pics}

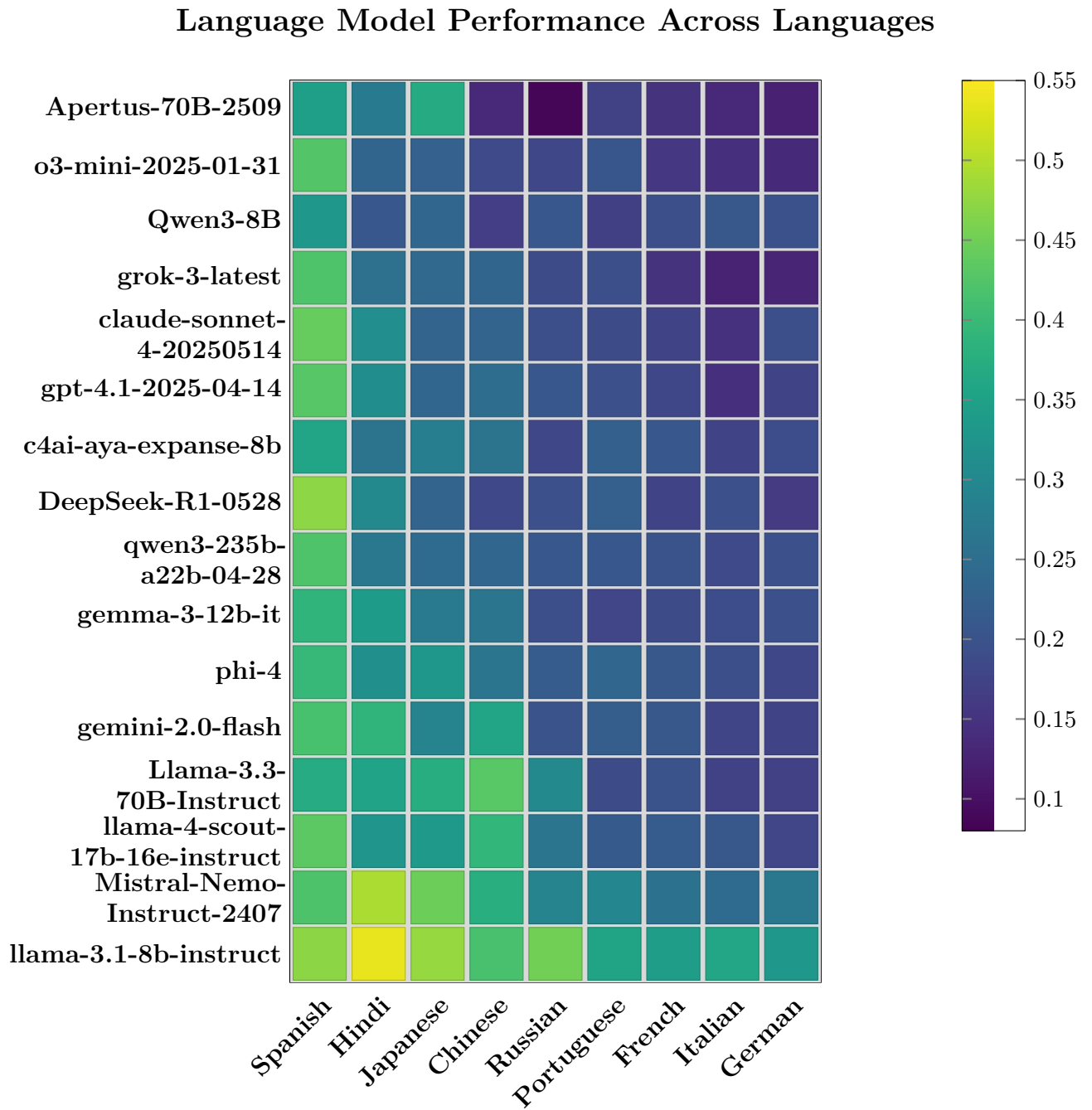
\begin{figure}[H]
    \caption{Spilling Heatmap of Different Language Models Across Various Languages. Scale indicate spilling rate.}
    \label{fig:language-model-heatmap}
\centering
\begin{tikzpicture}
\begin{axis}[
    width=10cm,
    height=16cm,
    title={Language Model Performance Across Languages},
    title style={font=\Large\bfseries, yshift=10pt},
    colorbar,
    colorbar style={
        ylabel={},
        ylabel style={rotate=-90, font=\large},
        width=1cm,
        height=12cm,
        samples=50,
    },
    colormap name=viridis,
    point meta min=0.08,
    point meta max=0.55,
    xmin=-0.5, xmax=8.5,
    ymin=-0.5, ymax=15.5,
    xtick={0,1,2,3,4,5,6,7,8},
    xticklabels={Spanish,	Hindi	,Japanese	,Chinese	,Russian	,Portuguese,	French,	Italian	,German
},
    xticklabel style={
        rotate=45, 
        anchor=north east, 
        font=\large\bfseries,
        text=black
    },
    xlabel={},
    xlabel style={font=\Large\bfseries, yshift=-15pt},
    ytick={0,1,2,3,4,5,6,7,8,9,10,11,12,13,14,15},
    yticklabels={
llama-3.1-8b-instruct,
Mistral-Nemo-Instruct-2407,
llama-4-scout-17b-16e-instruct,
Llama-3.3-70B-Instruct,
gemini-2.0-flash,
phi-4,
gemma-3-12b-it,
qwen3-235b-a22b-04-28,
DeepSeek-R1-0528,
c4ai-aya-expanse-8b,
gpt-4.1-2025-04-14,
claude-sonnet-4-20250514,
grok-3-latest,
Qwen3-8B,
o3-mini-2025-01-31,
Apertus-70B-2509
    },
    yticklabel style={
        font=\large\bfseries, 
        text width=4.5cm, 
        align=right,
        text=black
    },
    ylabel={},
    ylabel style={font=\Large\bfseries, xshift=-15pt},
    grid=none,
    axis equal=false,
    enlargelimits=false,
    axis line style={thick},
    tick style={thick},
    axis background/.style={fill=white},
]

\addplot[scatter, only marks, mark=square*, mark size=12pt, scatter src=explicit] 
table[x expr=0, y expr=\coordindex, meta=Spanish] \datatable;

\addplot[scatter, only marks, mark=square*, mark size=12pt, scatter src=explicit] 
table[x expr=1, y expr=\coordindex, meta=Hindi] \datatable;

\addplot[scatter, only marks, mark=square*, mark size=12pt, scatter src=explicit] 
table[x expr=2, y expr=\coordindex, meta=Japanese] \datatable;

\addplot[scatter, only marks, mark=square*, mark size=12pt, scatter src=explicit] 
table[x expr=3, y expr=\coordindex, meta=Chinese] \datatable;

\addplot[scatter, only marks, mark=square*, mark size=12pt, scatter src=explicit] 
table[x expr=4, y expr=\coordindex, meta=Russian] \datatable;

\addplot[scatter, only marks, mark=square*, mark size=12pt, scatter src=explicit] 
table[x expr=5, y expr=\coordindex, meta=Portuguese] \datatable;

\addplot[scatter, only marks, mark=square*, mark size=12pt, scatter src=explicit] 
table[x expr=6, y expr=\coordindex, meta=French] \datatable;

\addplot[scatter, only marks, mark=square*, mark size=12pt, scatter src=explicit] 
table[x expr=7, y expr=\coordindex, meta=Italian] \datatable;

\addplot[scatter, only marks, mark=square*, mark size=12pt, scatter src=explicit] 
table[x expr=8, y expr=\coordindex, meta=German] \datatable;

\draw[gray!30, thin] (axis cs:-0.5,{-0.5}) grid[step=1] (axis cs:8.5,15.5);

\end{axis}
\end{tikzpicture}
\end{figure}

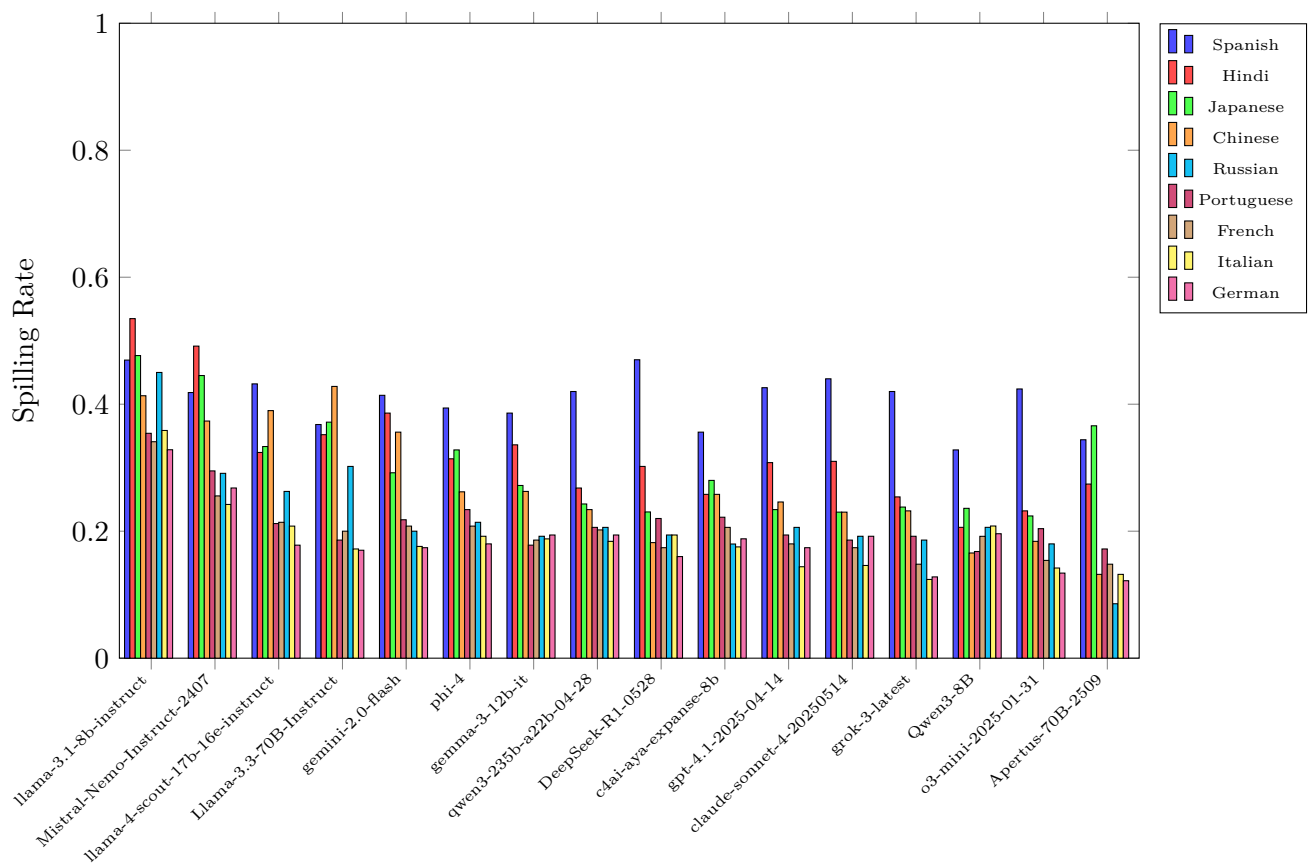
\begin{figure}[H]
\begin{tikzpicture}
\begin{axis}[
    ybar,
    width=15cm,
    height=10cm,
    ymin=0,
    ymax=1.0,
    xtick={0,1,2,3,4,5,6,7,8,9,10,11,12,13,14,15},
    xticklabels={
llama-3.1-8b-instruct,
Mistral-Nemo-Instruct-2407,
llama-4-scout-17b-16e-instruct,
Llama-3.3-70B-Instruct,
gemini-2.0-flash,
phi-4,
gemma-3-12b-it,
qwen3-235b-a22b-04-28,
DeepSeek-R1-0528,
c4ai-aya-expanse-8b,
gpt-4.1-2025-04-14,
claude-sonnet-4-20250514,
grok-3-latest,
Qwen3-8B,
o3-mini-2025-01-31,
Apertus-70B-2509    },
    xticklabel style={rotate=45, anchor=east, font=\tiny},
    xlabel style={yshift=-40pt},
    ylabel=Spilling Rate,
    legend style={at={(1.02,1)}, anchor=north west, font=\tiny},
    title={Detailed Histogram of Spilling Rate for Each Language and Model},
    title style={font=\large},
    bar width=2pt,
    enlarge x limits={abs=0.5},
]
\addplot[fill=blue!70, bar shift=-9pt] table[x expr=\coordindex, y=Spanish, col sep=comma] {data/sorted_language_model_performance.csv};
\addlegendentry{Spanish}
\addplot[fill=red!70, bar shift=-7pt] table[x expr=\coordindex, y=Hindi, col sep=comma] {data/sorted_language_model_performance.csv};
\addlegendentry{Hindi}
\addplot[fill=green!70, bar shift=-5pt] table[x expr=\coordindex, y=Japanese, col sep=comma] {data/sorted_language_model_performance.csv};
\addlegendentry{Japanese}
\addplot[fill=orange!70, bar shift=-3pt] table[x expr=\coordindex, y=Chinese, col sep=comma] {data/sorted_language_model_performance.csv};
\addlegendentry{Chinese}

\addplot[fill=cyan!70, bar shift=3pt] table[x expr=\coordindex, y=Russian, col sep=comma] {data/sorted_language_model_performance.csv};
\addlegendentry{Russian}
\addplot[fill=purple!70, bar shift=-1pt] table[x expr=\coordindex, y=Portuguese, col sep=comma] {data/sorted_language_model_performance.csv};
\addlegendentry{Portuguese}
\addplot[fill=brown!70, bar shift=1pt] table[x expr=\coordindex, y=French, col sep=comma] {data/sorted_language_model_performance.csv};
\addlegendentry{French}
\addplot[fill=yellow!70, bar shift=5pt] table[x expr=\coordindex, y=Italian, col sep=comma] {data/sorted_language_model_performance.csv};
\addlegendentry{Italian}
\addplot[fill=magenta!70, bar shift=7pt] table[x expr=\coordindex, y=German, col sep=comma] {data/sorted_language_model_performance.csv};
\addlegendentry{German}
\end{axis}
\end{tikzpicture}
\caption{Detailed Language Spilling Rate for Each Generative Model}
\label{fig:detailed_language_performance_results}
\end{figure}

In this section we provide a more detailed quantitative analysis of the spilling in models and languages. 
Figure \ref{fig:detailed_language_performance_results} presents a breakdown of the spilling rates for each generative model across the ten target languages. 
This allows for the identification of specific language pairs where language spilling is more or less pronounced for each model.

\section{Qualitative Analysis of the Responses}
\label{sec:apx_qualit}

\subsection{German}

\cherrypick{4o}{slip / Rutsch}{
    "Eindruck, Etwas Günstiges zu Erhalten": [
        "Er bekam den Job mit einem Rutsch.",
        "Die Beförderung kam mehr oder weniger mit einem Rutsch.",
        "Der Eintritt in die neue Firma verlief mit einem Rutsch."
    ],
}{
Using "Rutsch" for "getting something easily" is inocrrect. While "slip" in English can mean something obtained easily (like "it slipped through"), "Rutsch" in German doesn't carry this meaning. The idiomatic use doesn't transfer.  
}

\cherrypick{llama-70b}{charge / Ladung}{
  "Aufgabe oder Pflicht": [
    "Die Ladung des Chefs ist es, die Firma zu führen.",
    "Die Ladung des Lehrers ist es, die Schüler zu unterrichten.",
    "Die Ladung des Politikers ist es, das Wohl der Bürger zu vertreten."
  ]
}{
In English, "charge" can mean a responsibility or duty (e.g., "in charge of"), but "Ladung" in German does not carry this meaning. "Ladung" means electrical charge, cargo/freight, or explosive charge - not responsibility. German would use "Aufgabe," "Verantwortung," or "Pflicht" instead.
}

\subsection{French}

\cherrypick{4o}{collar / col}{
"Nœud ou raccord entre deux tronçons dans un tuyau": [
    "Le plombier a réparé le col qui fuyait dans le sous-sol.",
    "Il faut vérifier régulièrement le col des tuyaux pour éviter les fuites.",
    "L'installation comprenait plusieurs cols pour relier les différentes sections."
],
}{
The English definition of ``collar'' as a mechanical component (a restraining or connecting band, ring, or pipe) does not correspond to the standard usage of the French ``col'' for a pipe joint. This discrepancy indicates a potential instance where English vocabulary might be misleading.
}

\cherrypick{llama-8b}{tie / cravate}{
  "Cordelette" : [
    "Il a utilisé une cravate pour tirer la selle du cheval",
    "Les arbres étaient liés avec des cravates pour empêcher la chute des branches",
    "La cravate était trop serrée et a causé une douleur"
  ],
}{
Our initial word is cravate which means a necktie, a piece of clothing. 
However, it is used in a sense of a tie as a knot of a rope. The sentences translations are:
He used a [neck]tie to pull the horse saddle, The trees were tied with [neck]ties, The necktie was too tight and caused pain.
Only the last one can be used in that sense, but it seems like it interpreted the word tie in a sense of a knot.
}

\subsection{Italian}

\cherrypick{deepseek}{lift / ascensore}{
 "significato": ["un termine religioso o spirituale per indicare un' elevazione o un rinnovamento", ["L'ascensore durante la meditazione lo portò a una nuova consapevolezza.", "La sua fede lo ha guidato verso un ascensore spirituale.", "L'ascensore della sua anima lo portò a un livello di consapevolezza superiore."]]
}{
The use of ``ascensore'' as a metaphor for spiritual elevation is non-native and incorrect in Italian. The model appears to have mistakenly transferred a metaphor from English (such as the phrase ``lift spirits'') into Italian without considering idiomatic usage. Meanwhile, the spilling might come from French, from an expression ``être élevé spirituellement'' (to be spiritually lifted).
}

\cherrypick{4o}{cast / cast}{
    "gettare o lanciare qualcosa": [
        "Il pescatore ha cast la rete in mare.",
        "Ha cast il dado e sperava in un numero fortunato.",
        "Ha cast la palla al suo compagno di squadra."
    ],
}{
The English meaning of ``cast'' as in ``to cast a net'' is mistranslated in the Italian sentence. While in English, ``to cast a net'' is perfectly idiomatic and standard, this sense does not transfer directly into Italian using the verb ``cast''. In Italian, ``cast'' is a borrowed word that usually means the group of actors in a film, play, or show.
}

\subsection{Portuguese}

\cherrypick{llama-8b}{link / link}{
 "Relacionamento": [
    "O link entre eles era de amizade, mas agora é de rivalidade.",
    "O link entre a mãe e o filho é muito forte.",
    "O link entre as duas culturas é fundamental para o entendimento mútuo."
  ],
}{
In these 3 Portuguese examples, 3 models use link as a relationship, while the only meaning is a link on the internet, or in a document, which is an English influence. 
}

\cherrypick{llama-70b}{link / link}{
 "Relação": ["Há um link entre a dieta e a saúde", "O link entre a educação e o desenvolvimento econômico é bem estabelecido", "O link entre a poluição e as mudanças climáticas é um tema de debate"],
}{
}

\cherrypick{gemma-3}{link / link}{
 "relação (pessoas)": [
    "Ela sentiu um link imediato com o novo colega de trabalho.",
    "Eles compartilham um link profundo através de suas experiências em comum.",
    "Apesar da distância, eles mantiveram um link forte."
  ],
}{
}

\subsection{Spanish}

\cherrypick{llama-8b}{passage / pasaje}{
 "Intermedio o etapa en un proceso": [
    "La empresa tiene varios pasajes importantes en su historia.",
    "El pasaje de la vida de la empresa ha sido muy difícil.",
    "El pasaje de la infancia a la adolescencia es un momento crucial."
  ],
}{
Even though the word ``pasaje'' has a large variety of meanings, you cannot use this word as a step in a process. You can neither use passage as a step in English. The latent representations for passage and step must be close to each other. 
}

\cherrypick{gemini-small}{pack / paquete}{
 "Grupo de personas o cosas que actúan juntas": [
    "El equipo era un paquete de profesionales.",
    "El paquete de inversores estaba muy interesado en el proyecto.",
    "Era un paquete de amigos muy unidos."
  ],
}{
You cannot use a word ``paquete'' in Spanish meaning a group of people, but you can use this word in English denoting a group of wolves, or speaking informally a group of people that usually carries a negative emotion like ``pack of hooligans'', or ``A pack of reporters surrounded the celebrity''. 
}

\cherrypick{4o}{grip / agarre}{
 "Influencia o control sobre una situación": [
    "El nuevo director tiene un fuerte agarre sobre la empresa.",
    "Parece que la organización perdió el agarre sobre sus filiales en el extranjero.",
    "Ella siempre tuvo un agarre firme sobre su equipo, lo que le permitió liderar con éxito."
  ],
  "Entendimiento o comprensión de un concepto": [
    "Tiene un buen agarre de las teorías matemáticas avanzadas.",
    "Necesito mejorar mi agarre del idioma francés para poder comunicarme mejor.",
    "Su agarre sobre la historia medieval le hizo destacar en la clase."
  ],
}{
You cannot use a word \textit{agarre} in Spanish meaning the power, having a grip, influence, and control, but you can use a similar expression in English. 
At the same time, you can use a related expression in French, \textit{prendre en grippe}, which can be an additional spilling.
The second example uses the English meaning of a word \textit{grip} as (of a feeling or emotion) deeply affect (someone) like in a phrase ``she was gripped by a feeling of excitement'', and there is no such meaning of a word \textit{agarre} in Spanish
}
\subsection{Russian}
\cherrypick{anonymous}{collar / \cyrillicfont{ воротник}}{
\cyrillicfont{
  "значение": "Часть механизма или машины",
  "примеры": [
    "Воротник насоса был поврежден, и его нужно было заменить.",
    "Он регулировал воротник на шланге, чтобы вода текла медленнее.",
    "Воротник дробилки был забит, и ее нужно было прочистить."
  ]
}}{
The Russian word \ru{воротник} typically refers only to the part of clothing that surrounds the neck (e.g., shirt collar), and cannot be used to describe a mechanical component like a clamp or ring. However, in English, ``collar'' can mean a mechanical part (e.g., ``shaft collar'', ``pipe collar''), but this usage does not transfer into Russian, which is an example of spilling.
}

\cherrypick{llama-8b}{catch / \ru{улов}}{
\cyrillicfont{
  "уловляю": [
    "Я уловляю каждое слово, что говорит он.", 
    "Он уловлял каждое движение противника.", 
    "На эту игру я всегда уловляю."
  ],
}
}{
The usage of the word \textit{\ru{улов}} is incorrect in Russian to mean \textit{to understand}, which is used in this passage. The English verb \textit{catch} and synonyms \textit{to grasp} can be used to mean \textit{understand} and \textit{comprehend}. The noun \ru{улов} refers specifically to a \textit{catch of fish or animals}, not to comprehension or perception. This is a spilling.
}




\subsection{Chinese}

\cherrypick{claude}{match / {\cjkfont{火柴}}}{
{\cjkfont{ 
"体育比赛或竞赛": [
    "今晚有一场重要的足球火柴",
    "这场火柴的胜负关系到总冠军",
    "观众们都期待这场火柴的结果"
],}}
}{
The characters {\cjkfont{火柴}} mean only matches, or timber to start a fire. 
However, this example uses this word as a sport match, "Sports match or competition": ["There is an important football match tonight", "The outcome of this match is related to the championship", "The audience is looking forward to the result of this match"],
}

\cherrypick{deepseek-full}{mantle / {\cjkfont{地幔}}}{
{\cjkfont{"比喻责任或重担": [
    "他接过了领导团队的地幔，开始新的项目规划",
    "作为长子，他早早承担起家庭的地幔",
    "这份协议将环保地幔转移给了生产企业"
  ],}}
}{
The only meaning of  {\cjkfont{地幔}} is the Earth mantle layer, while in this example the model used the word as a meaning of this word from English, "an important role or responsibility that passes from one person to another", the translation is "Metaphor for responsibility or burden" with related examples. 
}

\subsection{Japanese}

\cherrypick{4o}{withdrawal / \japanesefont 引き出し}{
\japanesefont{ 
"5. 戦略的に後退または撤退すること": [
        "彼はすぐに話題から引き出した。",
        "部隊は安全のために戦線から引き出された。",
        "彼らは計画から引き出すことを決定した。"
    ]
}
}{
The Japanese word means only the drawer or an action of withdrawal, while the generated example means "The strategy of war means retreat and retreat" with the following examples, which is a direct spilling from an English word.
}

\cherrypick{4o-mini}{bank / \japanesefont 銀行}{
\japanesefont "川の土手や岸辺": (
"春になると、銀行に桜の花が咲きます。",
"子供たちは銀行で遊ぶのが大好きです。",
"銀行沿いの遊歩道を散歩しました。"
),
"データの保管場所": (
"このデータは専用の銀行に保存されています。",
"情報セキュリティのため、銀行は暗号化されています。",\\
"分析用のデータ銀行から情報を取得しました。"
),
}{
In this example the word means only the financial institution, while the model used meanings of the river bank, and data bank (term from informatics), the translations are "River banks and shorelines" and "data storage locations" respectively.
}
\subsection{Hindi}

\cherrypick{4o}{space / \hindifont अंतरिक्ष}{
\hindifont   "स्थान/जगह": [
        "कृपया अपनी किताबें इस अलमारी में व्यवस्थित करके रखें ताकि और अधिक अंतरिक्ष बच सके।",
        "हमारे घर में मेहमानों के लिए कोई अतिरिक्त अंतरिक्ष नहीं है।",
        "इस पार्किंग स्थल में आपकी गाड़ी के लिए पर्याप्त अंतरिक्ष है।"
    ],
     "अंतराल": [
        "संगीत के इस अंतरिक्ष के दौरान दर्शक तालियाँ बजाते हैं।",
        "इंटरव्यू के अंतरिक्ष में उम्मीदवारों ने अपने अनुभव साझा किए।",
        "अध्याय के विषयों के बीच एक अंतरिक्ष होना चाहिए ताकि पाठ समझने में आसानी हो।"
    "माध्यमिक क्षेत्र या स्थिति": [
        "उसके निर्णय ने उसे उसके करियर में एक नए अंतरिक्ष में पहुँचाया।",
        "इस वैश्विक समस्या को देखते हुए अब हमें पर्यावरण संरक्षण के एक नए अंतरिक्ष पर विचार करना होगा।",
        "समाज में महिलाओं की बढ़ती भूमिका एक सकारात्मक अंतरिक्ष में बदल रही है।"
    ]
 }
{
The first meaning refers to physical space — such as the space inside a room or ordinary spatial gaps.

The second meaning represents a gap or interval, which the English word space can also convey, but in Hindi it is expressed using words like \hi{अंतराल} or \hi{विराम}, not \hi{अंतरिक्ष}.

The third meaning refers to an abstract domain or sphere — for example, political space or creative space, meaning a new domain or area of activity.

In Hindi, however, the word \hi{अंतरिक्ष} refers only to cosmic space — the vast void of the universe beyond the Earth — and should not be used for these other meanings. In this example we can see how a single word produce a range of different meanings from English.
}

\section{Dictionary Sources and Validation}
\label{sec:dictionary_sources}

We employed authoritative lexicographic sources for each target language to provide robust reference frameworks for semantic validation:

\textbf{German:} DWDS (Digitales Wörterbuch der deutschen Sprache) - Digital Dictionary maintained by the Berlin-Brandenburg Academy of Sciences (\texttt{https://www.dwds.de/})

\textbf{French:} Larousse - Leading French lexicographic publisher with comprehensive coverage (\texttt{https://www.larousse.fr/})

\textbf{Portuguese:} Dicio - Comprehensive Portuguese dictionary (\texttt{https://www.dicio.com.br/})

\textbf{Spanish:} WordReference - Multilingual dictionary platform with extensive Spanish coverage (\texttt{http://wordreference.com/})

\textbf{English:} WordReference - Multilingual dictionary platform for English validation (\texttt{http://wordreference.com/})

\textbf{Italian:} Corriere Dizionari - Authoritative Italian dictionary by Corriere della Sera (\texttt{https://dizionari.corriere.it/})

\textbf{Chinese:} ZDIC (漢典) - Comprehensive Chinese character and word dictionary (\texttt{https://www.zdic.net/})

\textbf{Japanese:} Kotobank - Major Japanese dictionary aggregator (\texttt{https://kotobank.jp/})

\textbf{Russian:} Wiktionary Russian - Community-maintained multilingual dictionary (\texttt{https://ru.wiktionary.org/})

\textbf{Hindi:} Wiktionary Hindi - Comprehensive Hindi dictionary (\texttt{https://hi.wiktionary.org/})

These sources serve as principled baselines for comparative evaluation across multilingual models.

\section{Top-100 Highly Polysemous Words}

Our benchmark consists of the following 100 highly polysemous English words:

head, line, point, case, base, center, field, lead, stock, form, position, order, service, key, bar, plate, slip, life, hand, end, section, round, spot, pound, face, figure, job, hall, piece, stone, opening, band, twist, way, part, thing, house, side, rule, balance, division, release, defense, ball, extension, crown, stroke, shaft, man, number, body, control, book, voice, court, land, top, foot, issue, step, movement, bit, medium, game, union, sign, exchange, operation, resistance, corner, relief, frame, resolution, track, card, variation, score, split, strain, defence, wing, grain, draft, carrier, bull, pit, horn, pin, time, day, second, study, word, heart, view, action, front, king, force, review.

\section{Human Evaluation Instructions}
\label{sec:human_instructions}

The following instructions were provided to human evaluators for the semantic validation task. All participants provided informed consent for their data to be used for research purposes and model evaluation.

\subsection{Task 1: Semantic Validation}

\begin{enumerate}
\item Select the <language> and add the labels.
\item Mark each label as true if the meaning is correct, or false if it's not.
\item Provide the results as a string of t (true) and f (false), e.g., tfttf.
\item Access the evaluation interface at: \texttt{<url>}
\end{enumerate}

\subsection{Task 2: Dictionary Review}
\begin{enumerate}
\item Select the <language> and review the dictionary entries.
\item If you feel any definitions are incomplete or missing important meanings, you should add more.
\item You can check online dictionaries or other reliable sources to find additional meanings.
\item Copy and paste the additional meanings in the <language> from the online sources into the form.
\item Access the dictionary interface at: \texttt{<url>}
\end{enumerate}






    


\end{document}